\documentclass[11pt,a4paper]{article}
\usepackage[hyperref]{emnlp-ijcnlp-2019}
\usepackage{times}
\usepackage{latexsym}
\aclfinalcopy
\usepackage{url}

\usepackage{float}
\usepackage{dblfloatfix}
\usepackage{placeins}

\usepackage{booktabs} 

\usepackage{subfig}
\usepackage[toc,page]{appendix}
\usepackage{color}

\usepackage{amsmath}
\usepackage[utf8]{inputenc} 
\usepackage[T1]{fontenc}    
\usepackage{hyperref}       
\usepackage{url}            
\usepackage{booktabs}       
\usepackage{amsfonts}       
\usepackage{nicefrac}       
\usepackage{microtype}      
\usepackage{amsmath}
\usepackage{amsthm}
\usepackage{amssymb}
\usepackage{graphicx}

\usepackage{changes}
\usepackage{diagbox}

\definecolor{mygreen1}{rgb}{0, 0.5, 0.1}

\newcommand{\secmt}[1]{}

\newcommand{\ie}{\emph{i.e. }}

\definecolor{mypink1}{rgb}{0.858, 0.188, 0.478}

\definecolor{orange}{rgb}{1,0.5,0}

\definecolor{red}{rgb}{1,0,0}

\newcommand{\X}{\mathcal{X}}

\newcommand{\Y}{\mathcal{Y}}

\newcommand{\Tex}{\text{Tex}}
\newcommand{\Tok}{\text{Tok}}
\newcommand{\Tar}{\text{Tar}}

\newcommand{\Sent}{\text{Sent}}
\newcommand{\Pol}{\text{Pol}}

\makeatletter
\newcommand{\printfnsymbol}[1]{%
  \textsuperscript{\@fnsymbol{#1}}%
}

\title{From the \textit{Token} to the \textit{Review}: A Hierarchical Multimodal approach to Opinion Mining}

\author{Alexandre Garcia \thanks{\quad equal contribution}, Pierre Colombo\printfnsymbol{1}$^\dag$,  \\\textbf{Slim Essid, Florence d'Alch\'e-Buc, Chlo\'e Clavel}\\
  T\'el\'ecom ParisTech, Universit\'e Paris Saclay \\
  $\dag$ IBM GBS France\\
  \texttt{\{garcia,sessid,fdalche,cclavel\}@telecom-paristech.fr} \\
  \texttt{pierre.colombo@ibm.com}}

\date{}

\begin{document}
\maketitle

\begin{abstract}
The task of predicting fine grained user opinion based on spontaneous spoken language is a key problem arising in the development of Computational Agents as well as in the development of social network based opinion miners. Unfortunately, gathering reliable data on which a model can be trained is notoriously difficult and existing works rely only on coarsely labeled opinions. In this work we aim at bridging the gap separating fine grained opinion models already developed for written language and coarse grained models developed for spontaneous multimodal opinion mining. We take advantage of the implicit hierarchical structure of opinions to build a joint fine and coarse grained opinion model that exploits different views of the opinion expression. The resulting model shares some properties with attention-based models and is shown to provide competitive results on a recently released multimodal fine grained annotated corpus. 
\end{abstract}
\section{Introduction}

Recent years have witnessed the increasing popularity of social networks and video streaming platforms. 
People heavily rely on these channels to express their opinions through video-based discussions or reviews. Whereas such opinionated data has been widely studied in the context of written customer reviews \cite{liu2012sentiment} crawled on websites such as Amazon \cite{hu2004mining} and IMDB \cite{maas2011learning}, only a few studies have been proposed in the case of video-based reviews. 
Such multimodal data has been shown to provide a mean to disambiguate some hard to understand opinion expressions such as irony and sarcasm \cite{attardo2003multimodal} and contains crucial information indicating the level of engagement and the persuasiveness of the speaker \cite{clavel2016sentiment,ben2018,nojavanasghari2016deep}. A key problem in this context is the lack of availability of fine grained opinion annotation \ie annotations performed at the token or short span level and highlighting on the components of the structure of opinions. Indeed whereas such resources have been gathered in the case of textual data and can be used to deeply understand the expression of opinions \cite{wiebe2005annotating,pontiki2016semeval}, the different attempts at annotating multimodal reviews have shown that reaching good annotator agreement is nearly impossible at a fine grained level. This results from the disfluent aspect of spontaneous spoken language making it difficult to choose opinions' annotation boundaries \cite{garcia2019,langlet2015improving}. Thus the price to pay to gather reliable data is the definition of an annotation scheme focusing on coarse grained information such as long segment categorization as done by \citet{zadeh2016mosi} or review level annotation \cite{park2014computational}. Building models able to predict fine grained opinion information in a multimodal setting is in fact of high importance in the  context of designing human--robot interfaces \cite{langlet2016grounding}. Indeed the knowledge of opinions decomposed over a set of polarities associated to some targets is a building block of automatic human understanding pipelines \cite{langlet2015adapting}.
The present work is motivated by the following observations:
\begin{itemize}
    \item Despite the lack of reliability of fine grained labels collected for multimodal data, the redundancy of the opinion information contained at different granularities can be leveraged to reduce the inherent noise of the labelling process and to build improved opinion predictors. We build a model that takes advantage of this property and joinlty models the different components of an opinion.
    
    \item Hierarchical multi-task language models have been recently shown to improve upon the single tasks' models \cite{sanh2018hierarchical}. A careful choice of the tasks and the order in which they are sequentially presented to the model has been proved to be the key to build competitive predictors. It is not clear whether such type of hierarchical model could be adapted to handle multimodal data with the state of the art neural architectures \cite{zadeh2018multi,zadeh2018memory}. We discuss in the experimental section the strategies and models that are adapted to the multimodal opinion mining context.
    \item In the case where no fine grained supervision is available, the attention mechanism \cite{vaswani2017attention} provides a compelling alternative to build models generating interpretable decisions with token-level explanations \cite{hemamou2018hire}. In practice such models are notoriously hard to train and require the availability of very large datasets. On the other hand, the injection of fine-grained polarity information has been shown to be a key ingredient to build competitive sentiment predictors by \citet{socher2013recursive}. Our hierarchical approach can be interpreted under the lens of attention-based learning where some supervision is provided at training to counterbalance the difficulty of learning meaningful patterns with spoken language data. We specifically experimentally show that providing this supervision is here necessary to build competitive predictors due to the limited number of data and the difficulty to extract meaningful patterns from it.
\end{itemize}
\section{Background on fine grained opinion mining}
The computational models of opinion are grounded in a linguistic framework defining how these objects can be structured over a set of interdependent functional parts. In this work we focus on the model of \citet{martin2003language} that defines the expression of opinions as an \textit{evaluation} towards an object. The expression of such evaluations can be summarized by the combination of three components: a \textit{source} (mainly the speaker) expressing a statement on a \textit{target} identifying the entity evaluated and a \textit{polarized expression} making the attitude of the source explicit. In the literature, the task of finding the words indicating these components and categorizing them using a set of predefined possible targets and polarities has been studied under the name of Aspect Based Sentiment Analysis (ABSA) and popularized by the SEMEVAL campaigns \cite{pontiki2016semeval}. They defined a set of tasks including sentence-level prediction. \textit{Aspect Category Detection} consists in finding the target of an opinion from a set of possible entities; \textit{Opinion Target Expression} is a sequence tagging problem where the goal is to find the word indicating this entity; and \textit{Sentiment Polarity} recognition is a classification task where the predictor has to determine whether the underlying opinion is positive, negative or neutral. Such problems have also been extended at the text level (\textit{text-level ABSA}) where the participants were asked to predict a set of tuples (Entity category, Polarity level) summarizing the opinions contained in a review. In this work we adapt these tasks to a recently released fine-grained multimodal opinion mining corpus and study a category of hierarchical neural architecture able to jointly perform \textit{token}-level, \textit{sentence}-level and review-level predictions. In the next sections, we present the data available and the definition of the different tasks. 
\section{Data description and model}
\begin{figure*}[t]
    \centering
    \includegraphics[width=.9\textwidth]{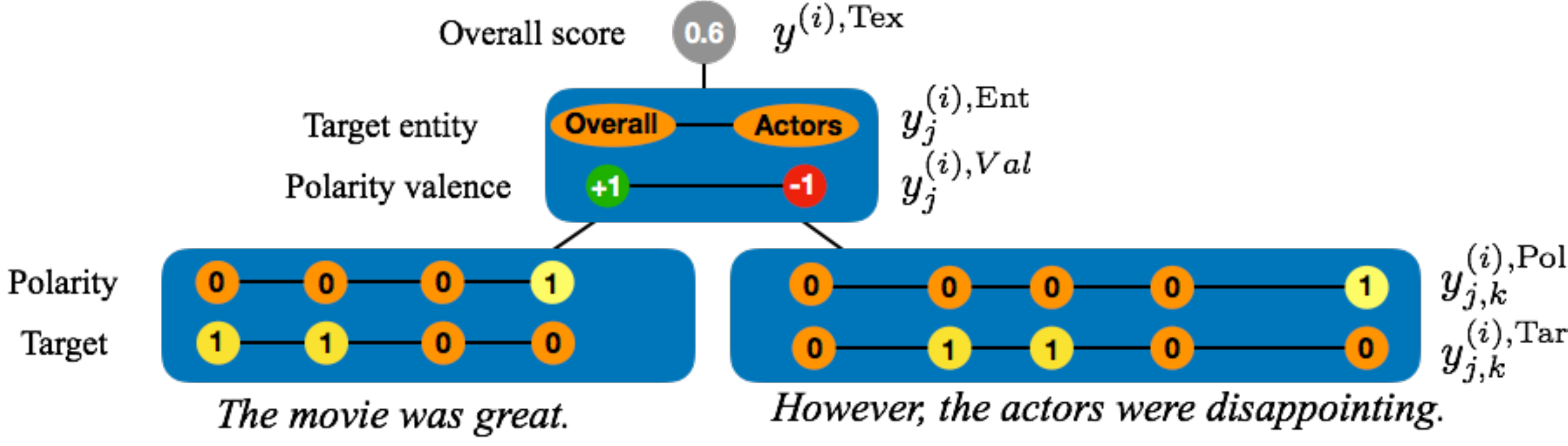}
    \caption{Structure of an annotated opinion}
    \label{Fig1}
\end{figure*}
This work relies on a set of fine and coarse grained opinion annotations gathered for the Persuasive Opinion Multimedia (POM) corpus presented in \citet{garcia2019}. The dataset is composed of 1000 videos carrying a strong opinion content: in each video, a single speaker in frontal view makes a critique of a movie that he/she has watched. The corpus contains 372 unique speakers and 600 unique movie titles. The opinion of each speaker has been annotated at 3 levels of granularity as shown in Figure \ref{Fig1}.

At the finest (\textit{Token}) level, the annotators indicated for each token whether it is responsible for the understanding of the polarity of the sentence and whether it describes the target of an opinion. 
On top of this, a span-level annotation contains a categorization of both the target and the polarity of the underlying opinion in a set of predefined possible target \textit{entities} and polarity \textit{valences}. 
At the review level (or \textit{text}-level since the annotations are aligned with the tokens of the transcript), an overall score describes the attitude of the reviewer about the movie.

As \citet{garcia2019} have shown that the boundaries of span-level annotations are unreliable, we relax the corresponding boundaries at the sentence level. This \textit{sentence} granularity is in our data the intermediate level of annotation between the \textit{token} and the \textit{text}. In practice, these intermediate level labels can be modeled by tuples such as the one provided in the \textit{text-level ABSA} SEMEVAL task which are given for each sentence in the dataset. In what follows, we will refer to the problem of predicting such information as the \textit{sentence level}-prediction problem. Details concerning the determination of the sentence boundaries and the associated pre-processing of the data are given in the supplemental material. 

The representation described above can be naturally converted into a mathematical representation: A review $\mathbf{x}^{(i)}, \; i\in \{1,\ldots,N\}$ is made of $S_i$ sentences each containing $W_{S_i}$ words. Thus the canonical feature representation of a review is the following $\mathbf{x}^{(i)} = \{\{x_{1,1}^{(i)},\ldots,x_{1,W_{S_1}}^{(i)}\},\ldots, \{x_{S_i,1}^{(i)},\ldots,x_{S_i,W_{S_i}}^{(i)}\} \}$, where each $x$ is the feature representation of a spoken word corresponding to the concatenation of a textual, audio and video feature representation. It has been shown in \cite{zadeh2018multi,zadeh2016mosi,zadeh2016multimodal} that whereas the textual modality carries the most information, taking into account video and audio modalities is mandatory to obtain state of the art results on sentiment analysis problems. Based on this input description, the learning task consists in finding a parameterized function $g_\theta:\X \rightarrow \Y$ that predicts various components of an opinion $\mathbf{y}\in \Y$ based on an input review $\mathbf{x}\in \X$. 
The parameters of such a function are obtained by minimizing an empirical risk:
\begin{equation}
    \hat{\theta} = \min_\theta \sum_{i = 1}^N l(g_\theta(\mathbf{x}^{(i)}),\mathbf{y}^{(i)}),
\end{equation}
where $l$ is a non-negative loss function penalizing wrong predictions. In general the loss $l$ is chosen as a surrogate of the evaluation metric whose purpose is to measure the similarity between the predictions and the true labels. In the case of complex objects such as opinions, there is no natural metric for measuring such proximity and we rely instead on distances defined on substructures of the opinion model. To introduce these distances, we first decompose the label-structures following the model previously described: 
\begin{itemize}
    \item  \textit{Token-level labels} are represented by a sequence of 2-dimensional binary label vectors $y^{(i),\Tok}_{j,k} = \begin{pmatrix} y^{(i),\Pol}_{j,k} \\ y^{(i),\Tar}_{j,k} \end{pmatrix}$ where $y^{(i),\Pol}_{j,k}$ and $y^{(i),\Tar}_{j,k}$ are some binary variables indicating respectively whether the $k^\text{th}$ word of the sentence $j$ in review $i$ is a word indicating the polarity of an opinion , and the target of an opinion.
    \item \textit{Sentence}-level labels carry 2 pieces of information: (1) the categorization of the target \textit{entities} mentioned in an opinion expressed is represented by an $E$ dimensional binary vector $y^{(i),\text{Ent}}_{j}$ where each component encodes the presence of an entity among $E$ possible values; and (2) the polarity of the opinions contained in the sentence are represented by a $4$-dimensional one-hot vector $y^{(i),\text{Val}}_{j}$ encoding the possible \textit{valences}: \textit{Positive}, \textit{Negative}, \textit{Neutral/Mixed} and \textit{None}. Thus the sentence level label $y^{(i),\Sent}_j$ is the concatenation of the two representations presented above:
    $y^{(i),\Sent}_j = \begin{pmatrix} y^{(i),\text{Ent}}_{j} \\ y^{(i),\text{Val}}_j \end{pmatrix}$ 
    \item \textit{Text}-level labels are composed of a single continuous score obtained for each review $y^{(i),Tex}$ summarizing the overall rating given by the reviewer to the movie described. 
\end{itemize}
Based on these representations, we define a set of losses, $l^{(\Tok)},l^{(\Sent)},l^{(\Tex)}$ dedicated to measuring the similarity of each substructure prediction, $\hat{\mathbf{y}}^{(\Tok)},\hat{\mathbf{y}}^{(\Sent)},\hat{\mathbf{y}}^{(\Tex)}$ with the ground-truth. In the case of binary variables and in the absence of prior preference between targets and polarities, we use the negative log-likelihood for each variable. Each task loss is then defined as the average of the negative log-likelihood computed on the variables that compose it. For continuous variables, we use the mean squared error as the task loss. Consequently the losses to minimize can be expressed as:
\begin{align*}
    &l^{(Tok)}(\mathbf{y}^{\Tok},\hat{\mathbf{y}}^{\Tok}) = - \frac{1}{2}\sum_i( \big(\mathbf{y}_i^{Pol}\log(\hat{\mathbf{y}}_i^{Pol}) + \\ & \quad \mathbf{y}_i^{Tar} \log( \hat{\mathbf{y}_i}^{Tar})\big),\\
    &l^{(Sent)}(\mathbf{y}^{Sent},\hat{\mathbf{y}}^{Sent}) = - \frac{1}{2} \sum_i \big(\mathbf{y}_i^{\text{Ent}}\log(\hat{\mathbf{y}}_i^{\text{Ent}}) + \\ &\quad \mathbf{y}_i^{\text{Val}}\log( \hat{\mathbf{y}}_i^{\text{Val}})\big),\\
    &l^{(Tex)}(y^{\Tex},\hat{y}^{\Tex}) = (y^{\Tex}-\hat{y}^{\Tex})^2,
\end{align*}

Following previous works on multi-task learning \cite{argyriou2007multi,ruder2017overview}, we argue that optimizing simultaneously the risks derived from these losses should improve the results, compared to the case where they are treated separately, due to the knowledge transferred across tasks. In the multi-task setting, the loss $l$ derived from a set of task losses $l^{(t)}$, is a convex combination of these different task losses. Here the tasks corresponds to each granularity level: $t\in \text{Tasks} = \{\textit{Tok},\textit{Sent},\textit{Tex}\}$ weighted according to a set of task weights $\lambda_t$ :
\begin{equation}
\label{eq:mtloss}
    l(\mathbf{y},\hat{\mathbf{y}}) = \frac{\sum_{t\in \text{Tasks}} \lambda_t l^{(t)}(\mathbf{y}^{t},\hat{\mathbf{y}}^{t})}{\sum_{t\in \text{Tasks}} \lambda_t},\; \forall \lambda_t \geq 0.
\end{equation}

Optimizing this type of objectives in the case of hierarchical deep net predictors requires building some strategy in order to train the different parts of the model: the low level parts as well as the abstract ones. We discuss such an issue in the next section.

\section{Learning strategies for multitask objectives}\label{sec:learning_strat}
The main concern when optimizing objectives of the form of \autoref{eq:mtloss} comes from the variable difficulty in optimizing the different objectives $l^{(t)}$. Previous works \cite{sanh2018hierarchical} have shown that a careful choice of the order in which they are introduced is a key ingredient to correctly train deep hierarchical models. In the case of hierarchical labels, a natural hierarchy in the prediction complexity is given by the problem. In the task at hand, coarse grained labels are predicted by taking advantage of the information coming from predicting fine grained ones. The model processes the text by recursively merging and selecting the information in order to build an abstract representation of the review. In Experiment 1 we show that incorporating these fine grained labels into the learning process is necessary to obtain competitive results from the resulting predictors. In order to gradually guide the model from easy tasks to harder ones, we parameterize each $\lambda_t$ as a function of the number of epochs of the form $\lambda^{(n_\text{epoch})}_t = \lambda_\text{max} \frac{\exp{((n_\text{epoch}-Ns_{t})/\sigma)}}{1+\exp{((n_\text{epoch}-Ns_{t})/\sigma)}}$ where $Ns_{t}$ is a parameter devoted to task $t$ controlling the number of epochs after which the weight switches to $\lambda_\text{max}$ and $\sigma$ is a parameter controlling the slope of the transition. We construct 4 strategies relying on smooth transitions from a low state $\lambda_t^{(0)}=0$ to a high state $\lambda_t^{(Ns_t)}=\lambda_t^{\max}$ of each task weight varying with the number of epochs. The different strategies described below are illustrated in the supplemental material.
\begin{itemize}
\item Strategy 1 (S1) consists in optimizing the different objectives one at a time from the easiest to the hardest. It consists in first moving vector $(\lambda_\textit{Token},\lambda_\textit{Sentence}, \lambda_\textit{Text})^T$ values from $( 1, 0, 0)^T$ to $(0, 1, 0)^T$ and then finally to $(0, 0, 1)^T$. The underlying idea is that the low level labels are only useful as an initialization point for higher level ones.
\item Strategy 2 (S2) consists in adding sequentially the different objectives to each other from the easiest to the hardest. It goes from a word only loss $(\lambda_\textit{Token}, \lambda_\textit{Sentence}, \lambda_\textit{Text})^T=(\lambda_\textit{Token}^{(N)}, 0, 0)^T$ and then adds the intermediate objectives by setting $\lambda_\textit{Sentence}$ to $\lambda_\textit{Sentence}^{(N)}$ and then $\lambda_\textit{Text}$ to $\lambda_\textit{Text}^{(N)}$. This strategy relies on the idea that keeping a supervision on low level labels has a regularizing effect on high level ones. Note that this strategy and the two following require a choice of the stationary weight values $\lambda_\textit{Token}^{(N)},\lambda_\textit{Sentence}^{(N)},\lambda_\textit{Text}^{(N)} $.
\item Strategy 3 (S3) is similar to (S2) except that the \textit{sentence} and \textit{text} weights are simultaneously increased. This strategy and the following one are introduced to test whether the order in which the tasks are introduced has some importance on the final scores.
\item Strategy 4 (S4) is also similar to (S2) except that \textit{text}-level supervision is introduced before the \textit{sentence}-level one. This strategy uses the intermediate level labels as a way to regularize the video level model that would have been learned directly after the \textit{token}-level supervision
\end{itemize}
These strategies can be implemented in any stochastic gradient training procedure of objectives (\autoref{eq:mtloss}) since it only requires modifying the values of the weight at the end of each epoch.
In the next section, we design a neural architecture that jointly predicts opinions at the three different levels, \textit{i.e.} the \textit{token}, \textit{sentence} and \textit{text} levels, and discuss how to optimize multitask objectives built on top of opinion-based output representations.
\section{Architecture}\label{sec:architecture}
Before digging into the model description, we introduce the set of hidden variables $h^{(i),\Tex},h^{(i),Sent}_j,h^{(i),\Tok}_{j,k}$ corresponding to the unconstrained scores used to predict the outputs: $\hat{y}^{(i),\Tex} = \sigma^\Tex(W^\Tex h^{(i),\Tex} + b^\Tex)$, $\hat{y}^{(i),\Sent}_j = \sigma^\Sent(W^\Sent h^{(i)^\Sent}_j+b^\Sent)$, $\hat{y}^{(i),\Tok}_{j,k}= \sigma^\Tok(W^\Tok h^{(i),\Tok}_{j,k} + b^\Tok)$, where the $W$ and $b$ are some parameters learned from data and the $\sigma$ are some fixed almost everywhere differentiable  functions ensuring that the outputs ``match'' the inputs of the loss function. In the case of binary variables for example, it is chosen as the sigmoid function $\sigma(x) = \exp(x)/(1+\exp(x))$. From a general perspective, a hierarchical opinion predictor is composed of 3 functions $g^\Tex,g^\Sent,g^\Tok$ encoding the dependency across the levels:
\begin{align*}
    h^{(i),\Tok}_{j,k} &= g^\Tok_{\theta^\Tok} ( x^{(i),\Tok}_{j,:}), \\
    h^{(i)^\Sent}_j &= g^\Sent_{\theta^\Sent} ( h^{(i)^\Tok}_{j,:}), \\
    h^{(i)^\Tex} &= g^\Tex_{\theta^\Tex} ( h^{(i)^\Sent}_{:}). 
\end{align*}
In this setting, low level hidden representations are shared with higher level ones. A large body of work has focused on the design of the $g$ functions in the case of multimodal inputs. In this work we exploit state of the art sequence encoders to build our hidden representations that we detail below. The mathematical expression of the models and a more in depth description are provided in the supplemental material.

\begin{itemize}
    \item Bidirectional Gated Recurrent Units (GRU) \cite{cho2014learning} especially when coupled with a self attention mechanism have been shown to provide state of the art results on tasks implying the encoding or decoding of a sentence in or from a fixed size representation. Such a problem is encountered in automatic machine translation \cite{luong2015effective}, automatic summarization \cite{nallapati2017summarunner} or image captioning and visual question answering \cite{anderson2018bottom}. We experiment with both models mixing the 3 concatenated input feature modalities (BiGRU model in Experiment 1) and a model carrying 3 independent BiGRU with a hidden state per modality (Ind BiGRU models).
    \item The Multi-attention Recurrent Network (MARN) proposed in \cite{zadeh2018multi} extends the traditional Long Short Term Memory (LSTM) sequential model by both storing a view specific dynamic (similar to the LSTM one) and by taking into account cross-view dynamics computed from the signal of the other modalities. In the original paper, this cross-view dynamic is computed using a multi-attention bloc containing a set of weights for each modality used to mix them in a joint hidden representation. Such a network can model complex dynamics but does not embed a mechanism dedicated to encoding very long-range dependencies.
    \item Memory Fusion Networks (MFN) are a second family of multi-view sequential models built upon a set of LSTM per modality feeding a joint delta memory. This architecture has been designed to carry some information in the memory even with very long sequences due to the choice of a complex retain / forget mechanism. 
\end{itemize}
The 3 models described previously build a hidden representation of the data contained in each sequence. The transfer from one level of the hierarchy to the next coarser one requires building a fixed length representation summarizing the sequence. Note that in the case of the MARN and the MFN, the model directly creates such a representation. We present the strategies that we deployed to pool these representations in the case of the BiGRU sequential layer.
\begin{itemize}
    \item Last state representation: Sequential models build their inner state based on observations from the past. One can thus naturally use the hidden state computed at the last observation of a sequence to represent the entire sequence. In our experiments, this is the representation chosen for the BiGRU and Ind BiGRU models.
    \item Attention based sequence summarization: Another technique consists in computing a weighted sum of the hidden states of the sequence. The attention weights can be learned from data to focus on the important parts of the sequence only and avoid building too complex inner representations. An example of such a technique successfully applied to the task of text classification based on 3 levels of representation can be found in \cite{yang2016hierarchical}. In our experiments, we implemented the attention model for predicting only the \textit{Sentence}-level labels (model Ind BiGRU + att Sent) and the \textit{Sentence} and \textit{Text}-level labels by sharing a common representation (Ind BiGRU + att model).
\end{itemize}
All the resulting architectures extend the existing hierarchical models by enabling the fusion of multimodal information at different granularity levels while maintaining the ability to introduce some supervision at any level.
\section{Experiments}
In this section we propose 3 sets of experiments that show the superiority of our model over existing approaches with respect to the difficulties highlighted in the introduction, and explore the question of the best way to train hierarchical models on multimodal opinion data.

All the results presented below have been obtained on the recently released fine grained annotated POM dataset \cite{garcia2019}. The input features are computed using the CMU-Multimodal SDK: We represented each word by the concatenation of the 3 feature modalities. The textual features are chosen as the 300-dimensional pre-trained Glove embeddings \cite{pennington2014glove} (not updated during training). The acoustic and visual features have been obtained by averaging the descriptors computed following \cite{park2014computational} during the time of pronunciation of each spoken word. These features include MFCC and pitch descriptors for the audio signals. For the video descriptors, posture, head and gaze movement are taken into account. As far as the output representations are concerned, we merely re-scaled the \textit{Text}-level polarity labels in the [0,1] range. 

The results are reported in terms of mean average error (MAE) for the continuous labels and micro F1 score $\mu F1$ for binary labels. We used the provided train, val and test set and describe for each experiment the training procedure and displayed values below. More detail concerning the preprocessings and architectures can be found in the supplemental material.

\subsection{Experiment 1: Which architecture provides the best results on the task of fine grained opinion polarity prediction?}
\begin{figure*}[hbt!]
    \centering
    \includegraphics[width=.8\textwidth]{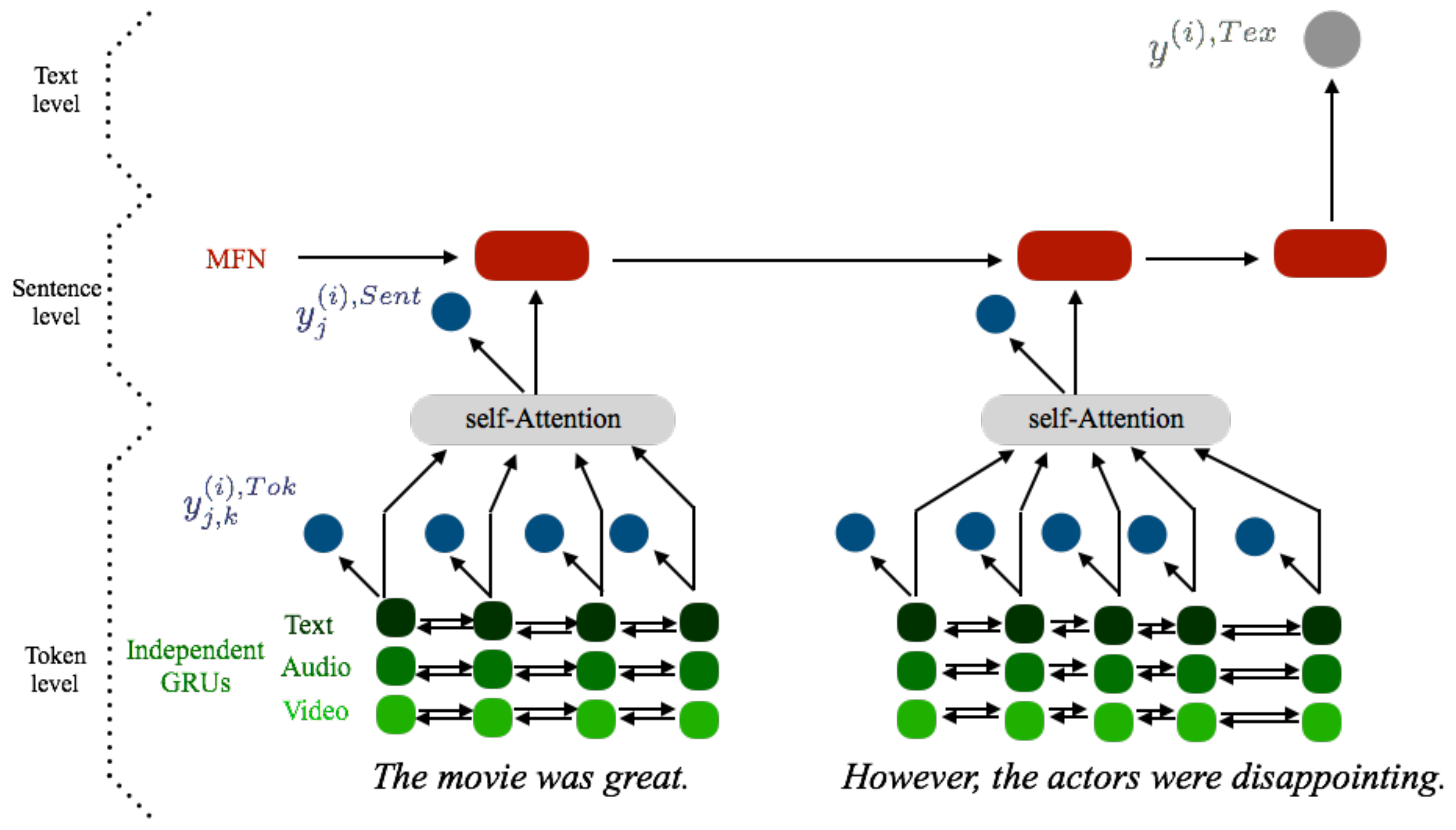}
    \caption{Best architecture selected during the Experiment 1}
    \label{Fig2}
\end{figure*}

In this first section, we describe our protocol to select an architecture devoted to performing fine grained multimodal opinion prediction. In order to focus on a restricted set of possible models, we only treat the polarity prediction problem in this section and selected the architectures that provided the best review-level scores (\ie with lowest mean average prediction error). Taking into account the entity categories would only bring an additional level of complexity that is not necessary in this first model selection phase. Building upon previous works \cite{zadeh2018memory}, we use the MFN model as our \textit{sentence}-level sequential model since it has been shown to provide state of the art results on \textit{text}-level prediction problems on the POM dataset. For the \textit{token}-level model, we test different state of the art models able to take advantage of the multimodal information. Our architecture is built upon the \textit{token}-level encoders presented in section \ref{sec:architecture}: the MFN, MARN and independent BiGRUs. Our baseline is computed similarly to \citet{zadeh2018multi}: we represent each sentence by taking the average of the feature representation of the Tokens composing it. The best results reported were obtained after a random search on the parameters and presented in Table \ref{tab:Exp1}.
\begin{table*}[hbt!]
\centering
\scalebox{0.9}{
\begin{tabular}{c|ccccccc}

\hline 
&\multicolumn{6}{c}{$\lambda_\textit{Tok}=\lambda_\textit{Sent}=0$: no fine grained supervision} \\\hline
MAE \textit{Text}    & 0.35 & 0.40 & 0.40& 0.38 &0.29 & 0.32 & \textbf{0.17}\\ \specialrule{.1em}{.0em}{0em}

&\multicolumn{6}{c}{Supervision at the token, sentence and review levels} \\\hline
\diagbox[height=20pt]{\footnotesize{Metric}}{\footnotesize{Model}} & {BiGRU} &{Ind BiGRU} & {Ind BiGRU + att Sent}& {Ind BiGRU + att}& {MARN}& {MFN}& {Av Emb}\\ \hline
$\mu F1$ \textit{Tokens}    & 0.90 & \textbf{0.93}  &
\textbf{0.93} & \textbf{0.93}
 & 0.90  & 0.89 & X\\ \hline
$\mu F1$ \textit{Sentence}    & 0.68 & 0.72 & \textbf{0.75}& \textbf{0.75} &0.52 & 0.47& X \\ \hline
MAE \textit{Text}    & 0.16 & 0.15 & 0.15& \textbf{0.14} &0.35 & 0.37 & X\\ \hline
\end{tabular}
}
\caption{Scores on sentiment label} 
\label{tab:Exp1}

\end{table*}
In the top row, we report results obtained when only using the \textit{text}-level labels to train the entire network. The baseline consisting in representing each sentence by the average of its tokens representation strongly outperforms all the other results. This is due to the moderate size of the training set (600 videos) which is not enough to learn meaningful fine grained representations. In the second part, we introduce some supervision at all levels and found that a choice of $\lambda_\textit{Tok}=0.05,\;\lambda_\textit{Sent}=0.5,\;\lambda_\textit{Tex} = 1$ being respectively the \textit{token}, \textit{sentence} and \textit{text} weights provides the best \textit{text}-level results. This combination reflects the fact that the main objective (\textit{text}-level) should receive the highest weight but low level ones also add some useful side supervision. Despite the ability of MARN and MFN to learn complex representations, the simpler BiGRU-based Token encoder retrieves the best results at all the levels and provides more than 12\% of relative improvement over the Average Embedding based model at the video level. This behavior reveals that the high complexity of MARN and MFN makes them hard to train in the context of hierarchical models with limited data leading to suboptimal performance against simpler ones such as BiGRU. We fix the best architecture obtained in this experiment displayed in Figure \ref{Fig2} and reuse it in the subsequent experiments. 
\subsection{Experiment 2: What is the best strategy to take into account multiple levels of opinion information?}
\begin{figure}[H]
    \centering
    \includegraphics[width=.5\textwidth]{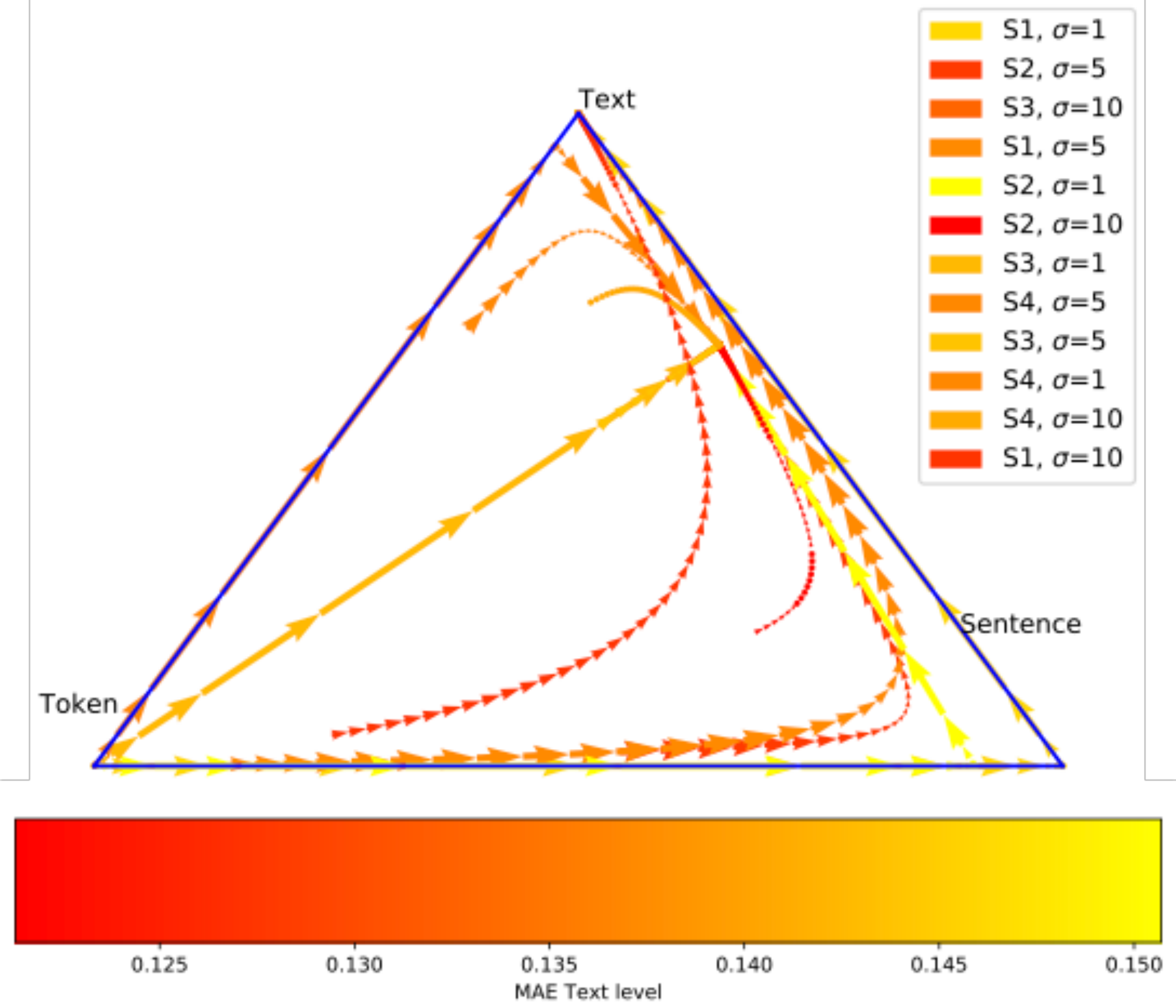}
    \caption{Path of the weight vector in the simplex triangle for the different tested strategies}
    \label{Fig3}
\end{figure}
Motivated by the issues concerning the training of multitask losses raised in Section \ref{sec:learning_strat}, we implemented the 4 strategies described and chose the final stationary values as the best one obtained in Experiment 1: $(\lambda_\textit{Token}^{(N)},\lambda_\textit{Sentence}^{(N)},\lambda_\textit{Text}^{(N)})= (0.05,0.5,1) $ Note that each strategy corresponds to a path of the vector $(\lambda_\textit{Tok},\lambda_\textit{Sent},\lambda_\textit{Tex})^T / \sum_t \lambda_t$ in the 3 dimensional simplex. We represent the 3 strategies tested in the Figure \ref{Fig3} corresponding to the projection of the weight vector onto the hyperplane containing the simplex. 

The best paths for optimizing the \textit{text}-level objectives are the one that smoothly move from a combination of \textit{sentence} and \textit{token}-level objectives to a \textit{text} oriented one. The path in the simplex seems to be more important than the nature of the strategy since S1 and S2 reach the same \textit{text}-level MAE score while working differently. It also appears than an objective with low $\sigma$\footnote{described in Section \ref{sec:learning_strat}} values corresponding to harder transitions tends to obtain lower scores than smooth transition based strategies. All the strategies are displayed as a function of the number of epochs in the supplemental material.
In this last section we deal with the issue of the joint prediction of entities and polarities.

\subsection{Experiment 3: Is it better to jointly predict opinions and entities ?}
In this section, we introduce the problem of predicting the entities of the movie on which the predictions are expressed, as well as the tokens that mention them. This task is harder than the previously studied polarity prediction task due to (1) the problem of label imbalance appearing in the label distribution reported in the Table \ref{Tab:4} and (2) the diversity of the vocabulary incurred when dealing with many entities. However since the presence of a polarity implies the presence of at least one entity, we expect that a joint prediction will perform better than an entitiy-based predictor only. Table \ref{Tab:3} contains the results obtained with the architecture described in Figure \ref{Fig2} on the task of joint polarity and entity prediction as well as the results obtained when dealing with these tasks independently. 

Using either the joint or the independent models provides the same results on the polarity prediction problems at the \textit{token} and \textit{sentence}-level. The reason is that the polarity prediction problem is easier and relying on the entities prediction would only introduce some noise in the prediction. 
\begin{table}[hbt!]
\begin{tabular}{c|ccc}
\hline
\multicolumn{1}{c|}{}                                              & \multicolumn{1}{c}{\begin{tabular}[c]{@{}c@{}}Polarity\\ labels\end{tabular}} & \begin{tabular}[c]{@{}c@{}}Entity\\ labels\end{tabular} & \begin{tabular}[c]{@{}c@{}}Polarity +\\ entities\end{tabular} \\ \hline
\begin{tabular}[c]{@{}c@{}}F1 polarity\\ tokens\end{tabular}  & 0.93  & X  & 0.93  \\ \hline
\begin{tabular}[c]{@{}c@{}}F1 polarity\\ valence\end{tabular} & 0.75   & X & 0.75 \\ \hline
\begin{tabular}[c]{@{}c@{}}F1 entities\\  tokens\end{tabular}  & X & 0.97 & 0.97 \\ \hline
\begin{tabular}[c]{@{}c@{}}F1 entities \\ Entities\end{tabular}  & X & Table \ref{Tab:4} & Table \ref{Tab:4} \\ \hline
\begin{tabular}[c]{@{}l@{}}MAE score \\ review level\end{tabular}   & 0.14 & 0.38 & 0.14  \\ \hline
\end{tabular}
\caption{Joint and independent prediction of entities and polarities}
\label{Tab:3}
\end{table}
We detail the case of \textit{Entities} in the Table \ref{Tab:4} and present the results obtained for the most common entity categories (among 11). As expected, the entity prediction tasks benefits from the polarity information on most of the categories except for the \textit{Vision and special effects}. A 5\% of relative improvement can be noted on the two most present \textit{Entities}: \textit{Overall} and \textit{Screenplay}. 
\begin{table}[hbt!]
\centering
\begin{tabular}{c|ccc}
\hline
 & Entity & \begin{tabular}[c]{@{}c@{}}Entity +\\ Polarity\end{tabular} & \begin{tabular}[c]{@{}c@{}}Value\\ Count\end{tabular} \\ \hline
Overall  & 0.71   & \textbf{0.73}  & 1985 \\ \hline
Actors   & \textbf{0.65}   & \textbf{0.65}    &  493   \\ \hline
Screenplay  & 0.60   & \textbf{0.63}  &  246\\ \hline
\begin{tabular}[c]{@{}l@{}}Atmosphere\\ and mood\end{tabular}   & 0.62   & \textbf{0.64}   &  151   \\ \hline
\begin{tabular}[c]{@{}l@{}}Vision and \\ special effects\end{tabular} & \textbf{0.62}   & 0.58  &  154      \\ \hline
\end{tabular}
\caption{F1 score per label for the top entity categories annotated at the sentence level (mean score averaged over 7 runs), value counts are provided on the test set.}
\label{Tab:4}
\end{table}
\section{Conclusion}
The proposed framework enables the joint prediction of the different components of an opinion based on a hierarchical neural network. The resulting models can be fully or partially supervised and take advantage of the information provided by different views of the opinions. We have experimentally shown that a good learning strategy should first rely on the easy tasks (\ie for which the labels do not require a complex transformation of the inputs) and then move to more abstract tasks by benefiting from the low level knowledge. Future work will explore the use of \textit{structured output learning} methods dedicated to the opinion structure.

\section{Acknowledgements}
We would like to thanks Thibaud Besson and the whole french Cognitive Systems team of IBM for supporting our research with the server IBM Power AC922.
\bibliography{acl2019}
\bibliographystyle{acl_natbib}

\pagebreak
\clearpage
\pagebreak

\appendix\section{Supplemental Material for the paper: From the \textit{Token} to the \textit{Review}: A joint Hierarchical Multimodal approach to Opinion Mining}
\label{sec:supplemental}

\subsection{Preprocessing details}

$\bullet$ \textit{Matching features and annotations:} In all our experiments we reused the descriptors presented originally in \cite{park2014computational} and made available in the CMU-Multimodal SDK. The annotation campaign performed in \cite{garcia2019} had been run on the unprocessed transcripts of the spoken reviews. In order to match the setting described in previous work, we transposed the fine grained annotations from the unprocessed dataset to the processed one in the following way: We first computed the Levenstein distance (minimum number of insertion/deletion/replacement needed to transform a sequence of items into another) between the sequence of Tokens of the processed and unprocessed transcripts. Then we applied the sequence of transformations minimizing this distance on the sequence of annotation tags to build the equivalent sequence of annotation on the processed dataset.

$\bullet$ \textit{Long sentences treatment:} 
We first removed the punctuation (denoted by the 'sp' token in the provided featurized dataset) in order to limit the maximal sentence length in the dataset. For the remaining sentences exceding 50 tokens we also applied the following treatment: We ran the sentence splitter from the spaCy library. The resulting subsentences are then kept each time they are composed of more than 4 tokens (overwise the groups of 4 tokens were merged with the next subsentence).

$\bullet$ \textit{Input features cliping:} The provided feature alignment code retrieved some infinite values and impossible assignments. We clipped the values to the range [-30,30] and replaced impossible assignments by 0.

$\bullet$ \textit{Training, validation and test folds:} We used the original standard folds available at: \url{https://github.com/A2Zadeh/CMU-MultimodalSDK/blob/master/mmsdk/mmdatasdk/dataset/standard_datasets/POM/pom_std_folds.py}
\subsection{Architecture details}
In this section we detail the structure of the different architectures tested in Experiment 1. According to the notations of the paper, we detail especially how the hidden representations $h^{(i),\Tex},h^{(i),Sent}_j,h^{(i),\Tok}_{j,k}$ are computed in practice. 
\subsubsection{\textit{token}-level model}
$\bullet$ BiGRU based models (For the sake of simplicity we consider only one direction for the equations)

The hidden state of a Gated recurrent unit at time $t$: $h_t^j$ is computed based on the previous state $h_{t-1}^j$ and a new candidate state $\tilde{h}_{t-1}^j$:
$$h_t^j = (1-z_j^t)h_{t-1}^j + z_t^j \tilde{h}_{t-1}^j $$
Where $z_t^j$ is an update vector controlling how much the state is updated :
$$z_j^t = \sigma( W_z \mathbf{x}^t + U_z \mathbf{h}_{t-1})^j $$
The candidate state is computed by a simple recurrent unit with an additional reset gate $\mathbf{r}_t$:
$$\tilde{h}_{t}^j  = (\text{tanh}(W \mathbf{x}_t + U (\mathbf{r}_t)\odot \mathbf{h}_{t-1} ))^j $$
$\odot$ is the element wise product and $\mathbf{r}_t$ is defined by:
$$r_t^j = \sigma (W_r \mathbf{x}_t + U_r \mathbf{h}_{t-1})^j$$.
\begin{itemize}
    \item In the case of the BiGRU model of Table \ref{tab:Exp1}, the input objects $\mathbf{x}_t$ are the concatenation of the 3 feature representations: $\mathbf{x}_t = x_t^\text{textual} \oplus x_t^\text{audio} \oplus x_t^\text{visual}$
    \item In the case of the Ind BiGRU Model, 3 BiGRU recurrent models are trained independently on each input modality and the hidden representation shared with the next parts of the network is the concatenation of the 3 hidden states: $\mathbf{h}_t = h_t^\text{textual} \oplus h_t^\text{audio} \oplus h_t^\text{visual}$
\end{itemize}
For these models, an entire sentence is encoded thanks to the state computed at the last token of the sentence. In the case of the BiGRU Ind + Att model, an additional attention model is used: it first computes a score per token $u_t^j$ indicating its relative contribution: 
$$u_t^j = \text{tanh}(W_w h_t^j + b_w)$$
These scores are then rescaled as a probability distribution with a softmax over the entire sequence:
$$\alpha_t = \frac{\mathbf{h}_t^T \mathbf{u}_t}{\sum_{t_j} \mathbf{h}_{t_j}^T \mathbf{u}_{t_j}}$$
These weights are then used to pool the hidden state representations of the sequence in a fixed length vector:
$$\mathbf{h}_\text{Pool} = \sum_t \alpha_t \mathbf{h}_t$$
This last representation is then used to feed the sentence level recurrent model (here a Memory fusion network). Note that the attention model does not erase the information about the modality nature of each component of $\mathbf{h}_\text{Pool}$ so that it can be used with a model taking into account this nature.

$\bullet$ MARN model

The Multi-Attention Recurrent Network from \citet{zadeh2018multi} relies on 2 components: 
\begin{itemize}
    \item The Long Short Term Hybrid Memory is a LSTM model where the hidden state is the concatenation of a local hidden state (computed using the classical LSTM layer) and an external hidden state computed using a Multi Attention Block (MAB).
    \item The MAB  block computes the external hidden state by computing 3 weighted sum of the input hidden states (one set of attention weights is computed per modality) which are then passed in 3 feedforward networks. The outputs of these network are concatenated and then passed in a second network to produce the final joint hidden representation. The detailed equations can be found in the original paper. 
\end{itemize}
Similarly to the original paper we use the last hidden state computed from an entire sequence to represent it.

$\bullet$ MFN model

The Memory Fusion Network \cite{zadeh2018memory} is made of 3 blocks:
\begin{itemize}
    \item Each modality based sequence of feature is represented by the hidden state of a LSTM model. These hidden state are fed in the next part of the model:
    \item A delta attention memory takes the concatenation of two consecutive input vectors (taken from the sequence of hidden representations of the LSTM) which are fed to a feedforward model to compute an attention score for each component of these inputs. The name delta memory is only indicating the fact that the inputs are taken by pairs of inputs.
    \item The output of the attention layer is then sent to a Multi-view Gated Memory generalizing the GRU layer to multiview data by taking into account a modality specific and a cross modality hidden representations. 
\end{itemize}

The MFN model is our common model at the \textit{sentence}-level.

\subsection{Hyperparameters}
All the hyper-parameters have been optimized on the validation set using MAE score at text level. Architecture optimization has been done using a random search with 15 trials. We used Adam optimizer \cite{adam} with a learning rate of 0.01, which is updated using a scheduler with a patience of 20 epochs and a decrease rate of 0.5 (one scheduler per classifier and per encoder). The gradient norm is clipped to 5.0, weight decay is set to 1e-5, and dropout \cite{dropout} is set to 0.2. Models have been implemented in PyTorch and they have been trained on a single IBM Power AC922.
The best performing MFN has a 4 attentions, the cellule size for the video is set to 48, for the audio to 32, for the text to 64. Memory dimension is set to 32, windows dimension to 2, hidden size of first attention is set to 32, hidden size of second attention is set to 16, $\gamma_1$ is set to 64, $\gamma_2$ is set to 32 \footnote{For exact meaning of each parameter please refer to the official implementation which can be found here: \url{https://github.com/pliang279/MFN} and in the work of \citet{zadeh2018memory}}.

\subsection{Experiment 2: Strategies displayed}

In this section we report the detailed expression of the $\lambda^{(n_\text{epoch})}$ displayed in the figure \ref{Fig3}.
\subsubsection{Strategy 1}

In the strategy 1, the task losses are activated one at a time following the equations:
\begin{align*}
    \lambda_\text{Token}^\text{nepoch} &=  1-\frac{\exp{((n_\text{epoch}-Ns_\text{Token})/\sigma)}}{1+\exp{((n_\text{epoch}-Ns_\text{Token})/\sigma)}} \\
    \lambda_\text{Sentence}^\text{nepoch} &=   \frac{\exp{((n_\text{epoch}-Ns_\text{Token})/\sigma)}}{1+\exp{((n_\text{epoch}-Ns_\text{Token})/\sigma)}} - \\ & \frac{\exp{((n_\text{epoch}-Ns_\text{Sentence})/\sigma)}}{1+\exp{((n_\text{epoch}-Ns_\text{Sentence})/\sigma)}}\\
    \lambda_\text{Text}^\text{nepoch} &= \frac{\exp{((n_\text{epoch}-Ns_\text{Sentence})/\sigma)}}{1+\exp{((n_\text{epoch}-Ns_\text{Sentence})/\sigma)}}
\end{align*}

\begin{figure}[ht]
    \centering
    \includegraphics[width=.45\textwidth]{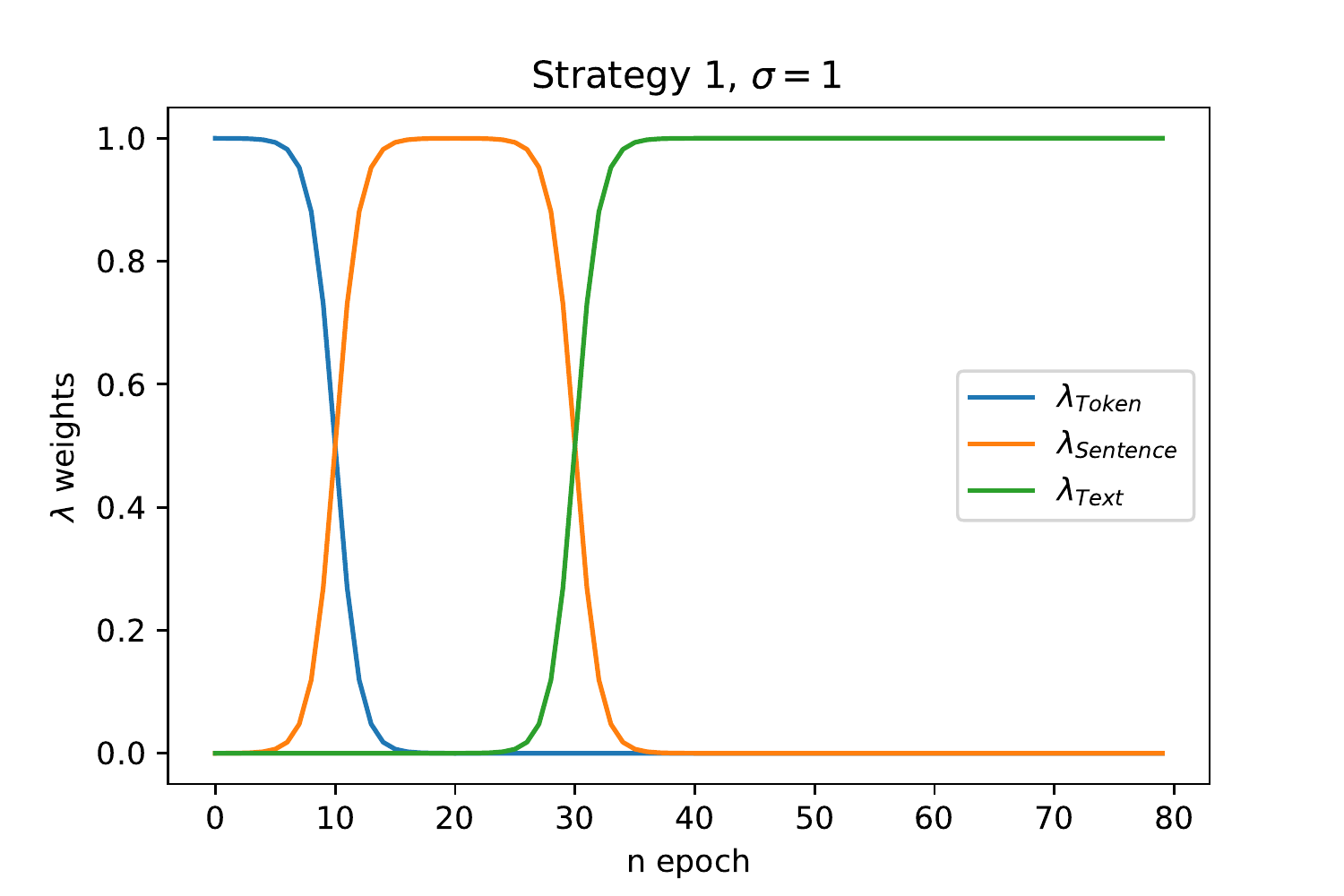}
    \caption{Strategy 1, $\sigma = 1$}
    \label{Strat1sig1}
\end{figure}

\begin{figure}[ht]
    \centering
    \includegraphics[width=.45\textwidth]{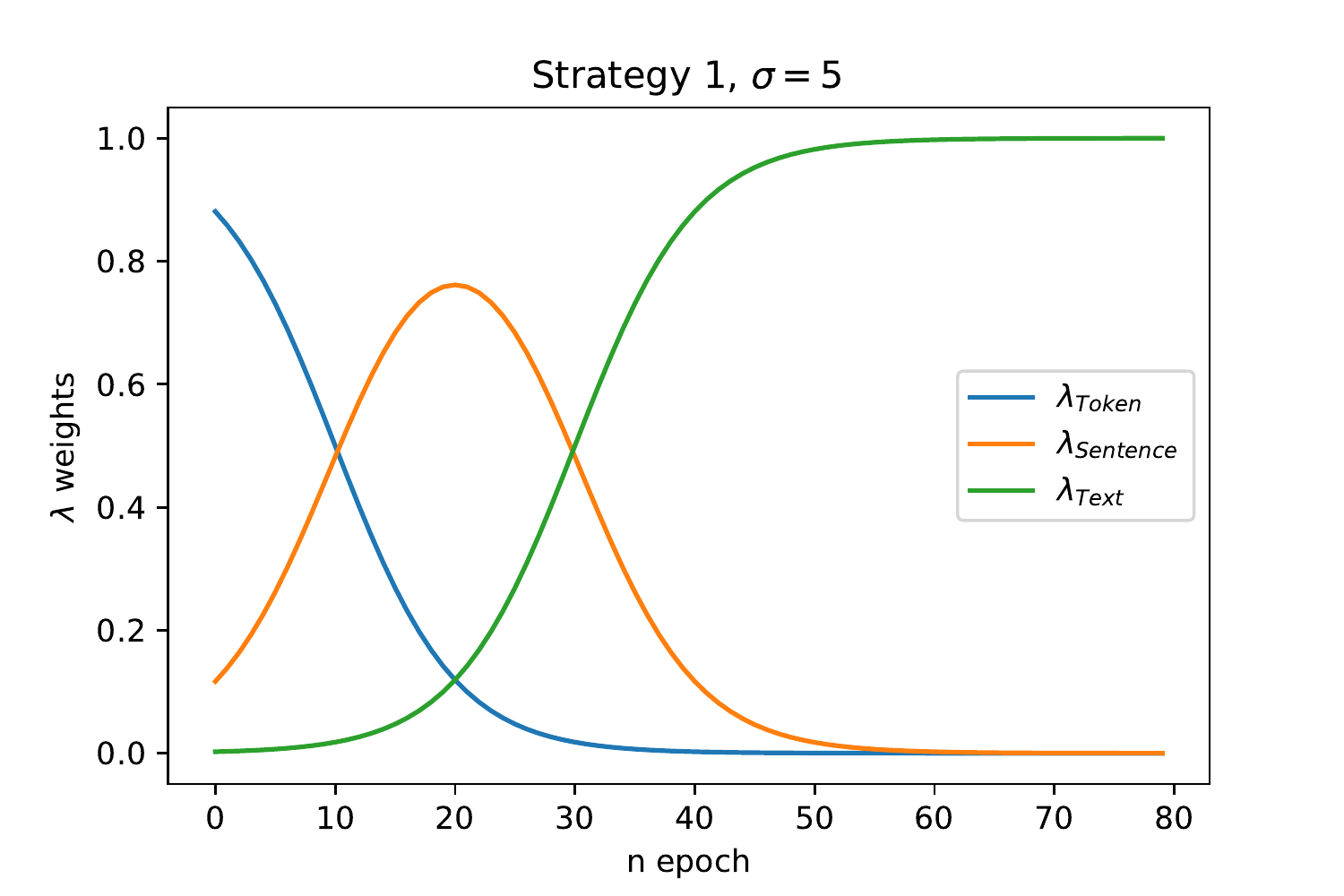}
    \caption{Strategy 1, $\sigma = 5$}
    \label{Strat1sig5}
\end{figure}

\begin{figure}[ht]
    \centering
    \includegraphics[width=.45\textwidth]{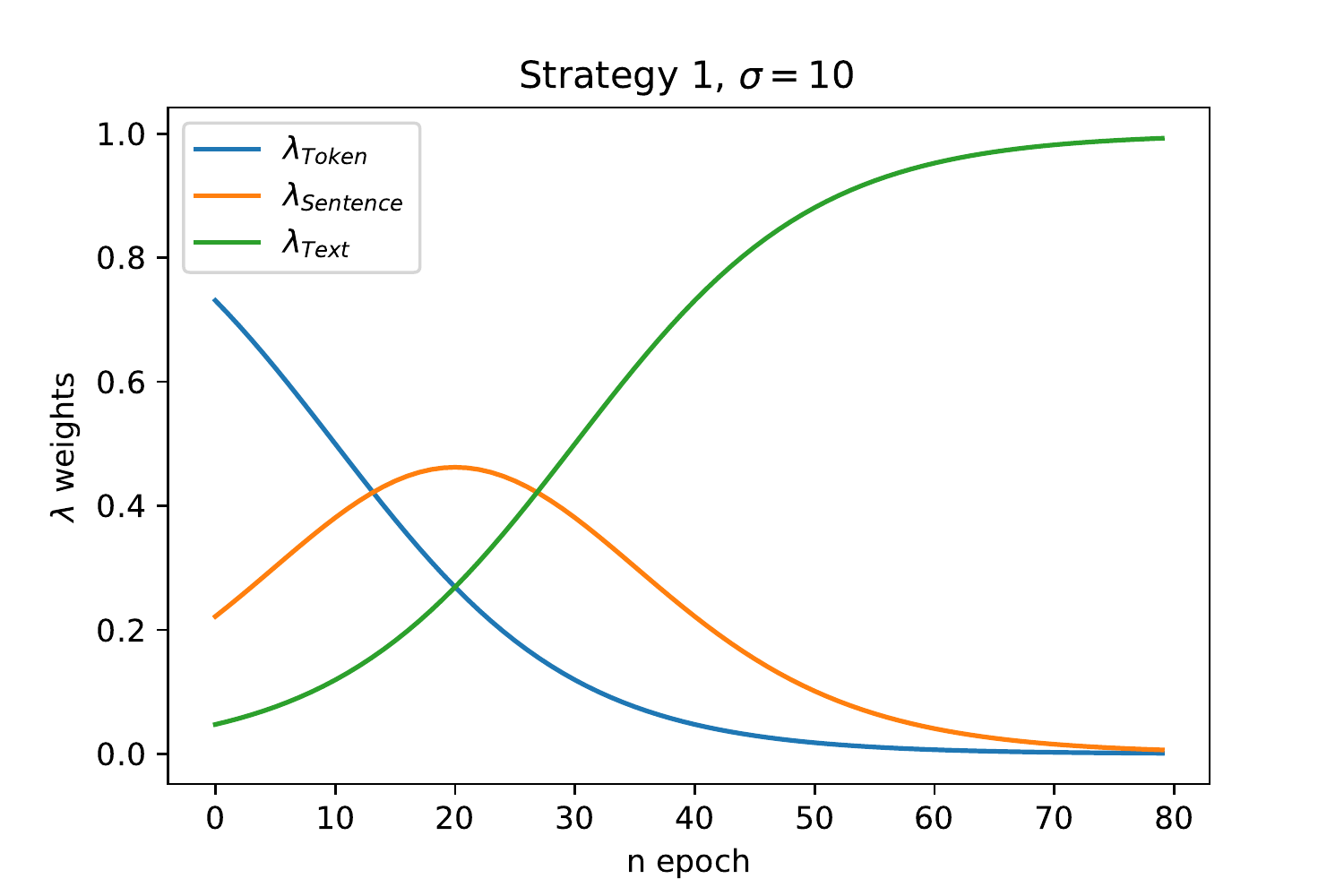}
   \caption{Strategy 1, $\sigma = 10$}
    \label{Strat1sig10}
\end{figure}

We report the graphs of the corresponding strategies as a function of the number of epochs in the Figures \ref{Strat1sig1},\ref{Strat1sig5} and \ref{Strat1sig10}.

\subsubsection{Strategy 2}

\begin{figure}[ht]
    \centering
    \includegraphics[width=.45\textwidth]{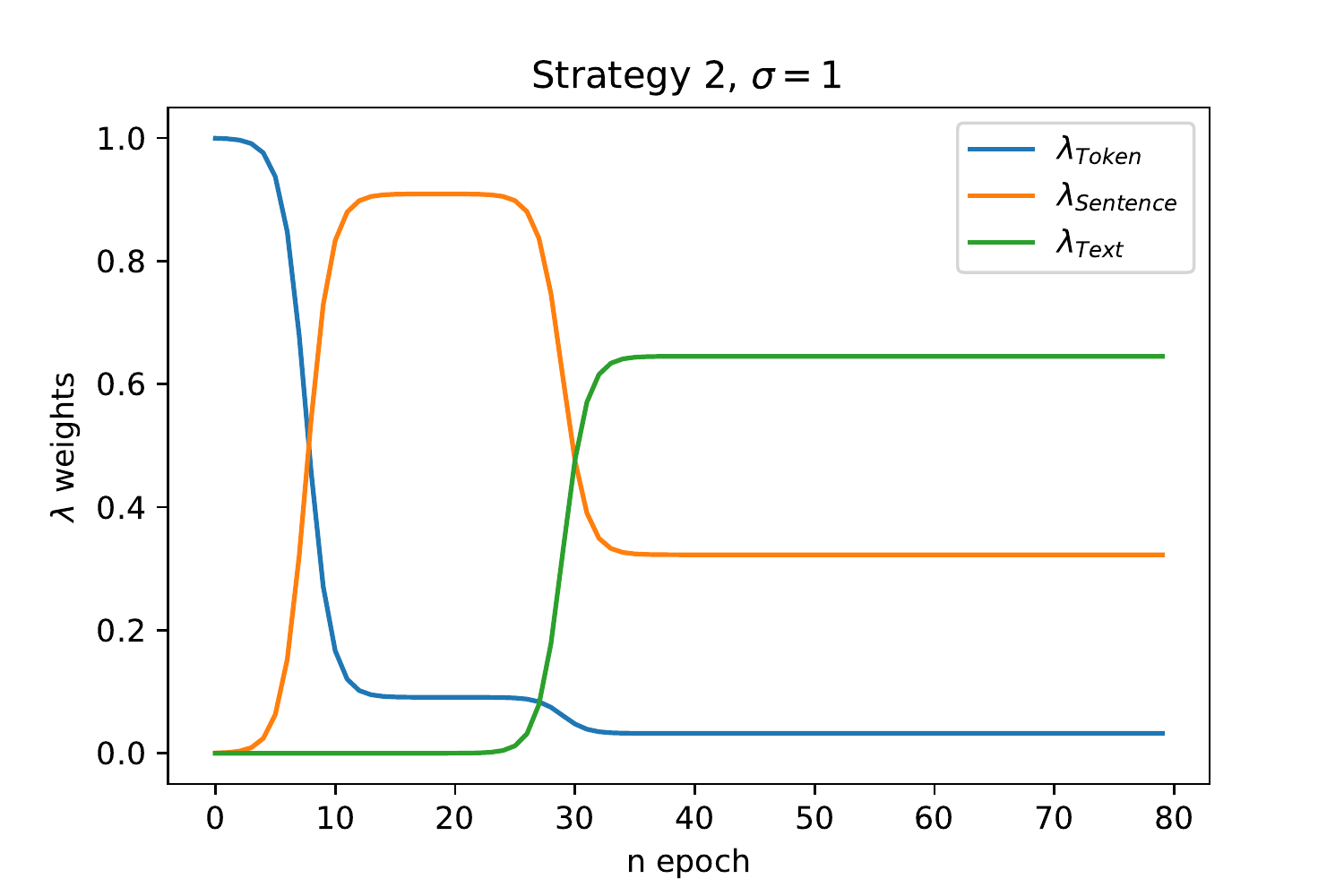}
    \caption{Strategy 2, $\sigma = 1$}
    \label{Strat2sig1}
\end{figure}

\begin{figure}[ht]
    \centering
    \includegraphics[width=.45\textwidth]{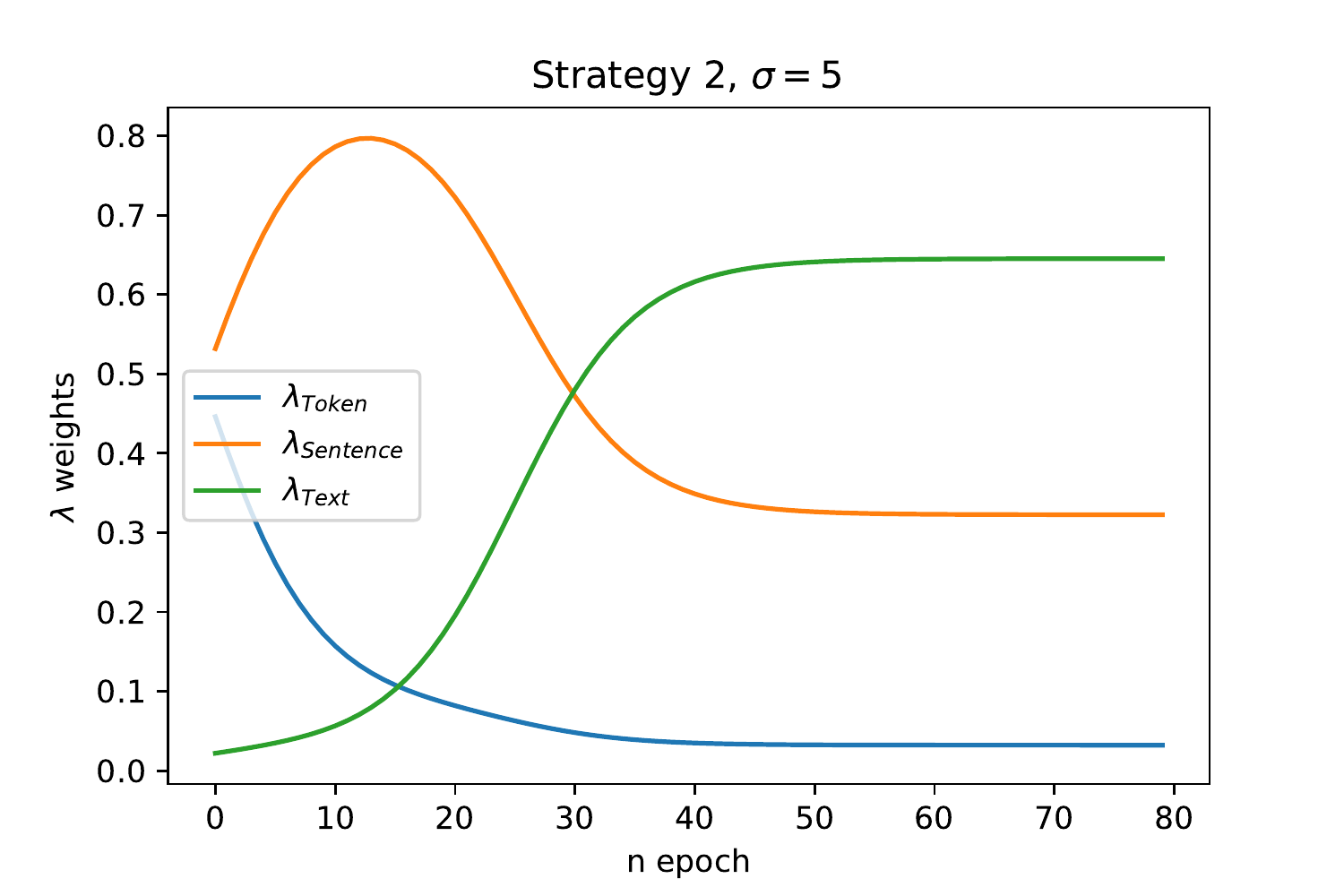}
    \caption{Strategy 2, $\sigma = 5$}
    \label{Strat2sig5}
\end{figure}

\begin{figure}[ht]
    \centering
    \includegraphics[width=.45\textwidth]{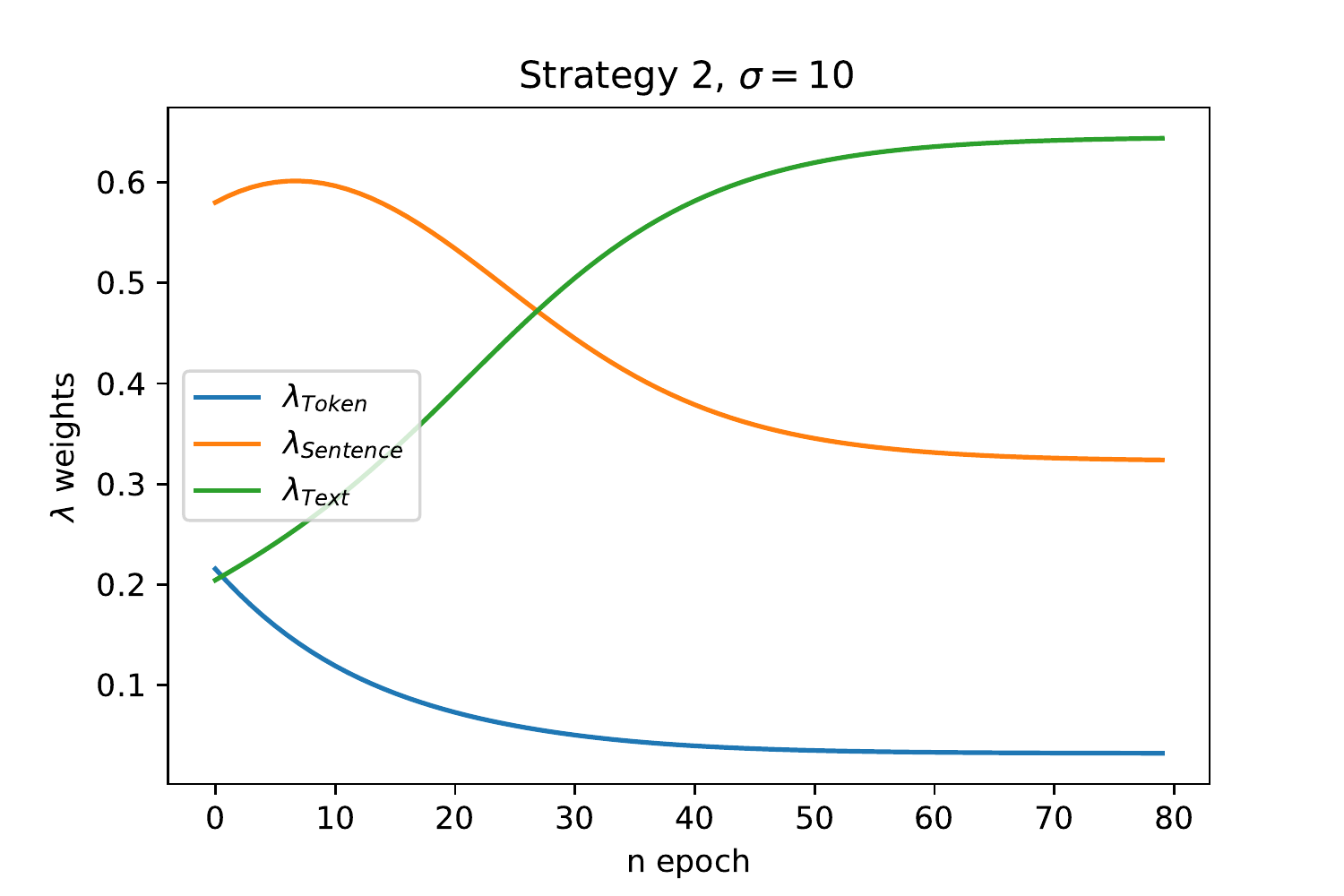}
    \caption{Strategy 2, $\sigma = 10$}
    \label{Strat2sig10}
\end{figure}

In the strategy 2, the task losses are sequentially activated and maintained following the equations:
\begin{align*}
    \lambda_\text{Token}^\text{nepoch} &=  0.05 \\
    \lambda_\text{Sentence}^\text{nepoch} &=  0.5 \frac{\exp{((n_\text{epoch}-Ns_\text{Token})/\sigma)}}{1+\exp{((n_\text{epoch}-Ns_\text{Token})/\sigma)}} \\
    \lambda_\text{Text}^\text{nepoch} &= \frac{\exp{((n_\text{epoch}-Ns_\text{Sentence})/\sigma)}}{1+\exp{((n_\text{epoch}-Ns_\text{Sentence})/\sigma)}}
\end{align*}
We report the graphs of the corresponding strategies as a function of the number of epochs in the Figures \ref{Strat2sig1},\ref{Strat2sig5} and \ref{Strat2sig10}.

\subsubsection{Strategy 3}

In the strategy 3, the \textit{Sentence} and \textit{Text} losses are activated at the same time:
\begin{align*}
    \lambda_\text{Token}^\text{nepoch} &=  0.05 \\
    \lambda_\text{Sentence}^\text{nepoch} &=  0.5 \frac{\exp{((n_\text{epoch}-Ns_\text{Token})/\sigma)}}{1+\exp{((n_\text{epoch}-Ns_\text{Token})/\sigma)}} \\
    \lambda_\text{Text}^\text{nepoch} &= \frac{\exp{((n_\text{epoch}-Ns_\text{Token})/\sigma)}}{1+\exp{((n_\text{epoch}-Ns_\text{Token})/\sigma)}}
\end{align*}

\begin{figure}[ht]
    \centering
    \includegraphics[width=.45\textwidth]{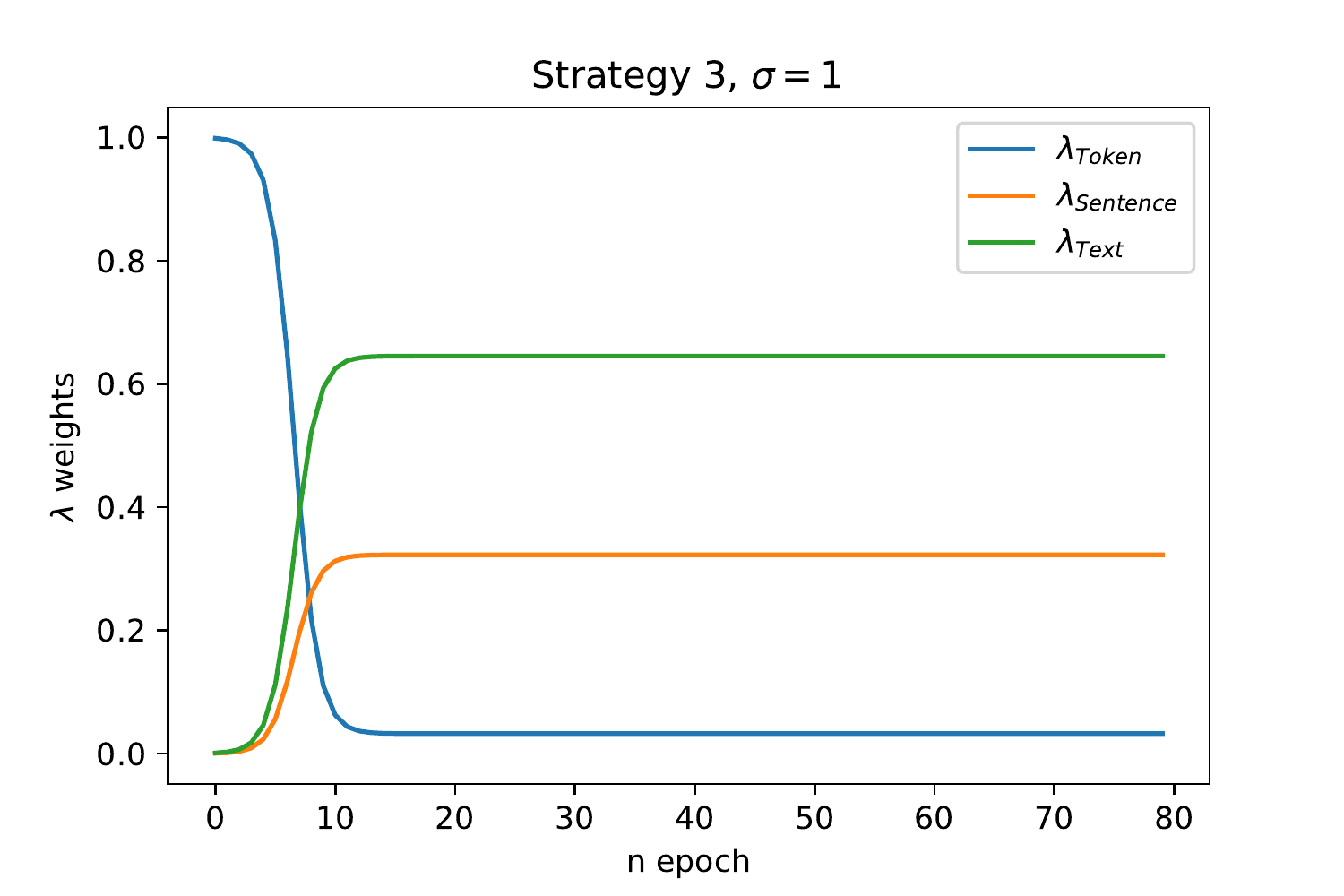}
    \caption{Strategy 3, $\sigma = 1$}
    \label{Strat3sig1}
\end{figure}

\begin{figure}[ht]
    \centering
    \includegraphics[width=.45\textwidth]{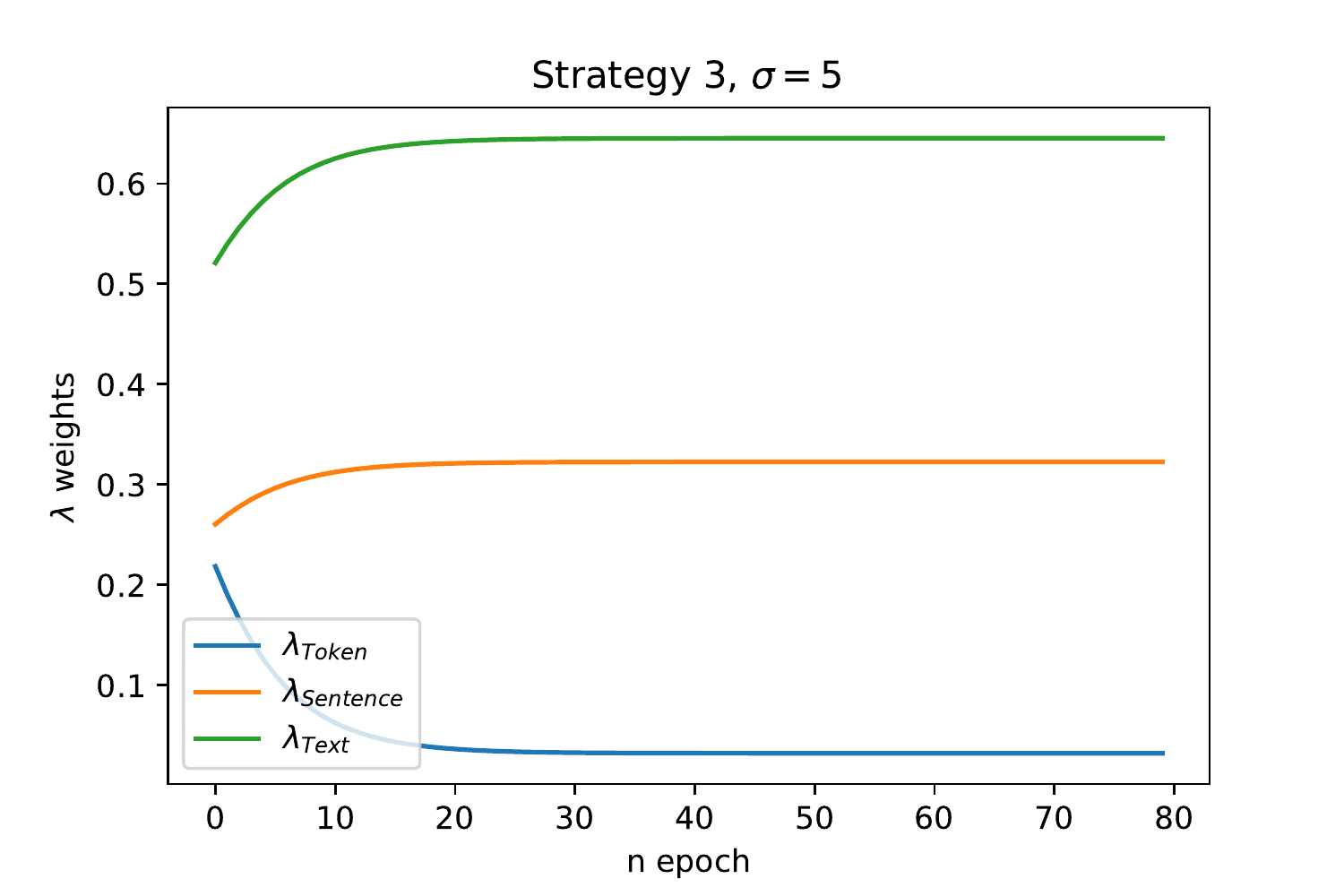}
    \caption{Strategy 3, $\sigma = 5$}
    \label{Strat3sig5}
\end{figure}

\begin{figure}[ht]
    \centering
    \includegraphics[width=.45\textwidth]{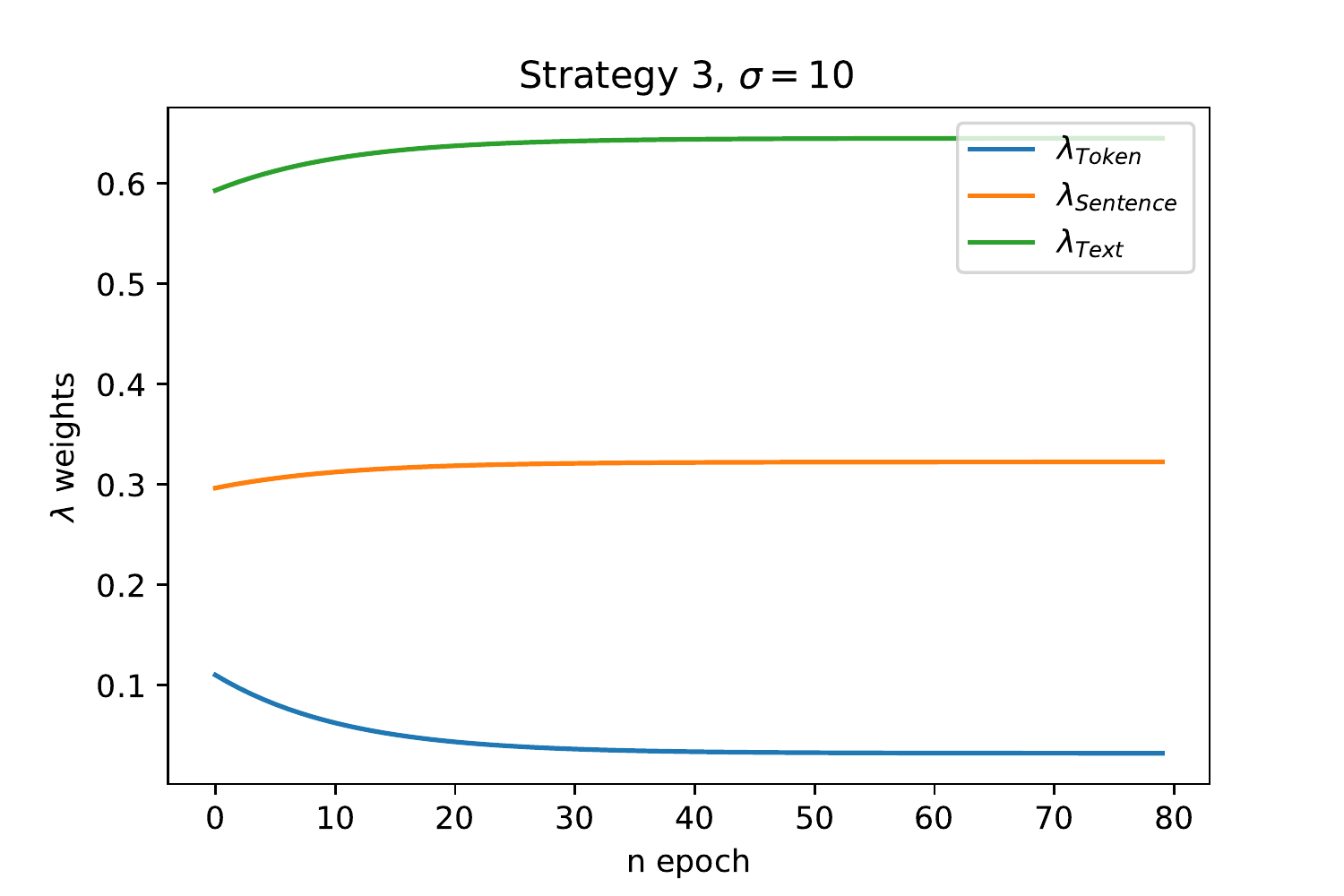}
    \caption{Strategy 3, $\sigma = 10$}
    \label{Strat3sig10}
\end{figure}

We report the graphs of the corresponding strategies as a function of the number of epochs in the Figures \ref{Strat3sig1},\ref{Strat3sig5} and \ref{Strat3sig10}.

\end{document}